\documentclass[journal]{IEEEtran}
\usepackage{tikz}
\newcommand\copyrighttext{%
  \footnotesize This paper is a preprint. IEEE copyright notice. ``\copyright~2019 IEEE. Personal use of this material is permitted. Permission from IEEE must be obtained for all other uses, in any current or future media, including reprinting/republishing this material for advertising or promotional purposes, creating new collective works, for resale or redistribution to servers or lists, or reuse of any copyrighted component of this work in other works.''}
\newcommand\copyrightnotice{%
\begin{tikzpicture}[remember picture,overlay]
\node[anchor=south,yshift=9pt] at (current page.south) {\fbox{\parbox{\dimexpr\textwidth-\fboxsep-\fboxrule\relax}{\copyrighttext}}};
\end{tikzpicture}
}

\usepackage{times}
\usepackage{epsfig}
\usepackage{graphicx}
\usepackage{amsmath}
\usepackage{amssymb}
\usepackage{array}
\usepackage{multirow}
\usepackage{subcaption}
\usepackage{color}
\usepackage{float}
\usepackage{graphicx}
\usepackage{mathtools}
\usepackage{booktabs}
\usepackage{hyperref}
\usepackage{mathtools}
\newcases{Rcases}{$\mskip\thickmuskip$}{%
  \hfil$\m@th{##}$}{$\m@th{##}$\hfil}{.}{\rbrace}

\usepackage[font=scriptsize]{caption}  
\begin{document}

\title{HybridSN: Exploring 3D-2D CNN Feature Hierarchy for Hyperspectral Image Classification}

\author{
Swalpa Kumar Roy,~\IEEEmembership{Student Member,~IEEE,}
Gopal Krishna,
Shiv Ram Dubey,~\IEEEmembership{Member,~IEEE,}
and
Bidyut B. Chaudhuri,~\IEEEmembership{Life~Fellow,~IEEE}
\thanks{S.K. Roy and G. Krishna are with Computer Science and Engineering Department at Jalpaiguri Government Engineering College, Jalpaiguri, West Bengal-735102, India (email: swalpa@cse.jgec.ac.in; gk1948@cse.jgec.ac.in).}
\thanks{S.R. Dubey is with Computer Vision Group, Indian Institute of Information Technology, Sri City, Chittoor, Andhra Pradesh-517646, India (e-mail: srdubey@iiits.in). }
\thanks{B.B. Chaudhuri is with Computer Vision and Pattern Recognition Unit at Indian Statistical Institute, Kolkata-700108, India (email: bbc@isical.ac.in).}
}

\markboth{Published in IEEE Geoscience and Remote Sensing Letters (DOI: \url{10.1109/LGRS.2019.2918719})}
 {Roy \MakeLowercase{\textit{et al.}}: Bare Demo of IEEEtran.cls for Journals}

\maketitle

\copyrightnotice

\begin{abstract}
Hyperspectral image (HSI) classification is widely used for the analysis of remotely sensed images. Hyperspectral imagery includes varying bands of  images. Convolutional Neural Network (CNN) is one of the most frequently used deep learning based methods for visual data processing.
The use of CNN for HSI classification is also visible in recent works. These approaches are mostly based on 2D CNN. Whereas, the HSI classification performance is highly dependent on both spatial and spectral information. Very few methods have utilized the 3D CNN because of increased computational complexity. 
This letter proposes a Hybrid Spectral Convolutional Neural Network (HybridSN) for HSI classification. Basically, the HybridSN is a spectral-spatial 3D-CNN followed by spatial 2D-CNN. The 3D-CNN facilitates the joint spatial-spectral feature representation from a stack of spectral bands. The 2D-CNN on top of the 3D-CNN further learns more abstract level spatial representation. Moreover, the use of hybrid CNNs reduces the complexity of the model compared to 3D-CNN alone. To test the performance of this hybrid approach, very rigorous HSI classification experiments are performed over Indian Pines, Pavia University and Salinas Scene remote sensing datasets. The results are compared with the state-of-the-art hand-crafted as well as end-to-end deep learning based methods. A very satisfactory performance is obtained using the proposed HybridSN for HSI classification. The source code can be found at \url{https://github.com/gokriznastic/HybridSN}.
\end{abstract}


\section{Introduction} 
\IEEEPARstart{T}{he} research in hyperspectral image analysis is important due to its potential applications in real life \cite{chang2003hyperspectral}. Hyperspectral imaging results in multiple bands of images which makes the analysis challenging due to increased volume of data. The spectral, as well as the spatial correlation between different bands convey useful information regarding the scene of interest. Recently, Camps-Valls et al. have surveyed the advances in hyperspectral image (HSI) classification \cite{camps2014advances}. The HSI classification is tackled in two ways, one with hand-designed feature extraction technique and another with learning based feature extraction technique.

Several HSI classification approaches have been developed using the hand-designed feature description \cite{yang2018hyperspectral,fang2018new}. Yang and Qian have proposed a  joint collaborative representation by using the locally adaptive dictionary \cite{yang2018hyperspectral}. It reduces the adverse impact of useless pixels and improves the HSI classification performance. Fang et al. have utilized the local covariance matrix to encode the relationship between different spectral bands \cite{fang2018new}. They used these matrices for HSI training and classification using Support Vector Machine (SVM). A composite kernel is used to combine spatial and spectral information for HSI classification \cite{camps2006composite}. Li et al. have applied the learning over the combination of multiple features for the classification of hyperspectral scenes \cite{li2015multiple}.
Some other hand crafted approaches are 
Joint Sparse Model and Discontinuity Preserving Relaxation \cite{gao2018hyperspectral}, Boltzmann Entropy-Based Band Selection \cite{gao2018boltzmann}, Sparse Self-Representation \cite{hu2018band}, Fusing Correlation Coefficient and Sparse Representation \cite{tu2018hyperspectral}, Multiscale Superpixels and Guided Filter \cite{dundar2018sparse}, and etc. 

\begin{figure*}[!t]
\centering
\includegraphics[clip=true, trim = 03 550 10 10, width=\textwidth]{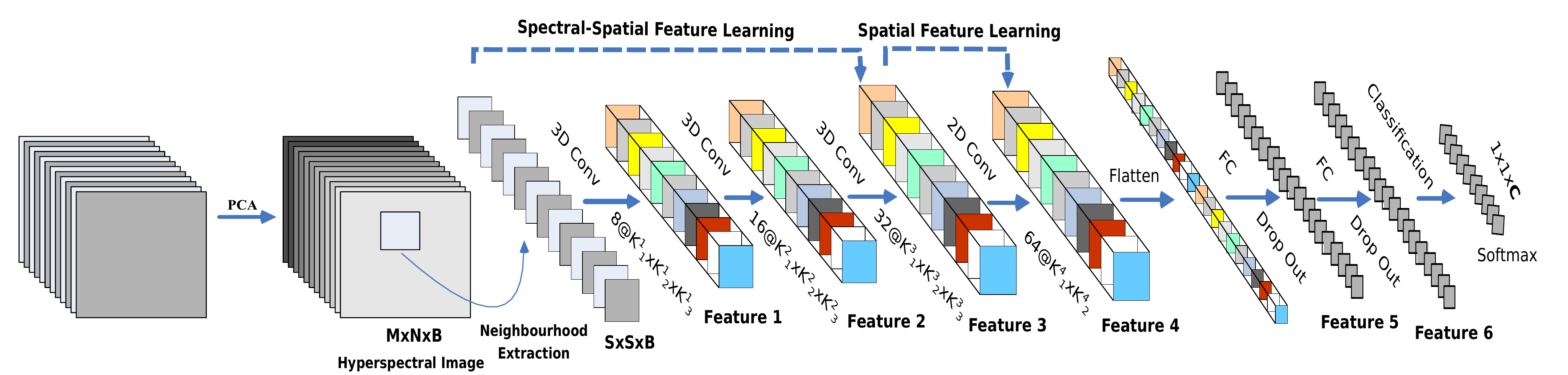}
\caption{Proposed HybridSpectralNet (HybridSN) Model which integrates 3D and 2D convolutions for hyperspectral image (HSI) classification.}
\label{fig:model}
\end{figure*}

Recently, the Convolutional Neural Network (CNN) has become very popular due to drastic performance gain over the hand-designed features \cite{krizhevsky2012imagenet}. The CNN has shown very promising performance in many applications where visual information processing is required, such as image classification \cite{he2016deep,roy2019lisht}, object detection \cite{ren2015faster}, semantic segmentation \cite{he2017mask}, colon cancer classification \cite{basha2018rccnet}, depth estimation \cite{repala2018dual}, face anti-spoofing \cite{nagpal2018performance}, etc.
In recent years, a huge progress is also made in deep learning for hyperspectral image analysis. A dual-path network (DPN) by combining the residual network and dense convolutional network is proposed for the HSI classification \cite{kang2018dual}. Yu et al. have proposed a greedy layer-wise approach for unsupervised training to represent the remote sensing images \cite{yu2018unsupervised}. Li et al. introduced a pixel-block pair (PBP) based data augmentation technique to generalize the deep learning for HSI classification \cite{li2018data}. Song et al. have proposed deep feature fusion network \cite{song2018deep} and Cheng et al. have used the off-the-shelf CNN models for HSI classification \cite{cheng2018exploring}. Basically, they extracted the hierarchical deep spatial features and used with SVM for training and classification. Recently, the low power consuming hardwares for deep learning based HSI classification is also explored \cite{haut2018low}. Chen et al. have used the deep feature extraction of 3D-CNN for HSI classification \cite{chen2016deep}. Zhong et al. have proposed the spectral-spatial residual network (SSRN) \cite{zhong2018spectral}. The residual blocks in SSRN use the identity mapping to connect every other 3-D convolutional layer. Mou et al. have investigated the residual conv-deconv network, an unsupervised model, for HSI classification \cite{mou2018unsupervised}. Recently, Paoletti et al. have proposed the Deep Pyramidal Residual Networks (DPRN) specially for the HSI data \cite{paoletti2018deep}. Very recently, Paoletti et al. have also proposed  spectral-spatial capsule networks to learn the hyperspectral features \cite{paoletti2018capsule}, whereas Fang et al. introduced deep hashing neural networks for hyperspectral image feature extraction~\cite{fang2019hashing}.

It is evident from the literature that using just 2D-CNN or 3D-CNN had a few shortcomings such as missing channel relationship information or very complex model, respectively. It also prevented these methods from achieving a better accuracy on hyperspectral images. The main reason is due to the fact that hyperspectral images are volumetric data and have a spectral dimension as well. The 2D-CNN alone isn't able to extract good discriminating feature maps from the spectral dimensions. Similarly, a deep 3D-CNN is more computationally complex and alone seems to perform worse for classes having similar textures over many spectral bands. This is the motivation for us to propose a hybrid-CNN model which overcomes these shortcomings of the previous models. The 3D-CNN and 2D-CNN layers are assembled for the proposed model in such a way that they utilise both the spectral as well as spatial feature maps to their full extent to achieve maximum possible accuracy.

This letter proposes the HybridSN in Section 2; presents the experiments and analysis in Section 3; and highlights the concluding remarks in section 4.

\section{Proposed HybridSN Model}
Let the spectral-spatial hyperspectral data cube be denoted by $\mathbf{I}\in\mathcal{R}^{M\times{N}\times{D}}$, where $\mathbf{I}$ is the original input, $M$ is the width, $N$ is the height, and $D$ is the number of spectral bands/depth. Every HSI pixel in $\mathbf{I}$ 
contains $D$ spectral measures and forms a one-hot label vector $Y = (y_1, y_2, \ldots y_{C}) \in \mathcal{R}^{1\times{1}\times{C}}$, where $C$ represents the land-cover categories. However, the hyperspectral pixels exhibit the mixed land-cover classes, introducing the high intra-class variability and inter-class similarity into $\mathbf{I}$. It is of great challenge for any model to tackle this problem. 
To remove the spectral redundancy first the traditional principal component analysis (PCA) is applied over the original HSI data ($\mathbf{I}$) along spectral bands. The PCA reduces the number of spectral bands from $D$ to $B$ while maintaining the same spatial dimensions (i.e., width $M$ and height $N$). We have reduced only spectral bands such that it preserves the spatial information which is very important for recognising any object. We represent the PCA reduced data cube by $\mathbf{X}\in\mathcal{R}^{M\times{N}\times{B}}$, where  $\mathbf{X}$ is the modified input after PCA, $M$ is the width, $N$ is the height, and $B$ is the number of spectral bands after PCA.

In order to utilize the image classification techniques, the HSI data cube is divided into small overlapping 3D-patches, the truth labels of which are decided by the label of the centred pixel. We have created the 3D neighboring patches $P\in\mathcal{R}^{S\times{S}\times{B}}$ from $\mathbf{X}$, centered at the spatial location $(\alpha,\beta)$, covering the $S\times S$ window or spatial extent and all $B$ spectral bands. The total number of generated 3D-patches ($n$) from $X$ is given by $(M-S+1) \times (N-S+1)$.
Thus, the 3D-patch at location $(\alpha,\beta)$, denoted by $P_{\alpha,\beta}$, covers the width from $\alpha - (S-1)/2$ to $\alpha + (S-1)/2$, height from $\beta - (S-1)/2$ to $\beta + (S-1)/2$ and all $B$ spectral bands of PCA reduced data cube $X$.

\begin{table}[!t]
\caption{The layer wise summary of the proposed $HybridSN$ architecture with window size 25$\times$25. The last layer is based on the Indian Pines dataset.}
\resizebox{\linewidth}{!}{
\begin{tabular}{ c c c }
 \toprule
 Layer (type) & Output Shape &  \# Parameter  \\ \midrule 
input\_1 (InputLayer) & (25, 25, 30, 1) & 0  \\  
conv3d\_1 (Conv3D) & (23, 23, 24, 8) & 512  \\
conv3d\_2 (Conv3D) & (21, 21, 20, 16) & 5776 \\
conv3d\_3 (Conv3D) & (19, 19, 18, 32) & 13856 \\
reshape\_1 (Reshape) & (19, 19, 576) & 0 \\
conv2d\_1 (Conv2D) & (17, 17, 64) & 331840 \\
flatten\_1 (Flatten) & (18496) & 0 \\
dense\_1 (Dense) & (256) & 4735232\\
dropout\_1 (Dropout) & (256) & 0 \\
dense\_2 (Dense) & (128) & 32896 \\
dropout\_2 (Dropout) & (128) & 0 \\
dense\_3 (Dense) & (16) & 2064 \\
\bottomrule
Total Trainable Parameters: $5,122,176$ \\
\bottomrule
\end{tabular}}
\label{tab:model_summ}
\end{table}


\begin{table*}[!t]
\caption{The classification accuracies (in percentages) on Indian Pines, University of Pavia, and Salinas Scene datasets using proposed and state-of-the-art methods.}
\centering
\begin{tabular}{c|ccc|ccc|ccc}
\toprule
\multirow{2}{*}{Methods} & \multicolumn{3}{c|}{Indian Pines Dataset} & \multicolumn{3}{c|}{University of Pavia Dataset} & \multicolumn{3}{c}{Salinas Scene Dataset}\\
\cline{2-10}
 & OA & Kappa & AA & OA & Kappa & AA & OA & Kappa & AA \\ 
\midrule
SVM & 85.30 $\pm$ 2.8 & 83.10 $\pm$ 3.2 & 79.03 $\pm$ 2.7 & 94.34 $\pm$ 0.2 & 92.50 $\pm$ 0.7 & 92.98 $\pm$ 0.4 & 92.95 $\pm$ 0.3 & 92.11 $\pm$ 0.2 & 94.60 $\pm$ 2.3\\
2D-CNN & 89.48 $\pm$ 0.2 & 87.96 $\pm$ 0.5 & 86.14 $\pm$ 0.8 & 97.86 $\pm$ 0.2 & 97.16 $\pm$ 0.5 & 96.55 $\pm$ 0.0 & 97.38 $\pm$ 0.0 & 97.08 $\pm$ 0.1 & 98.84 $\pm$ 0.1\\
3D-CNN & 91.10 $\pm$ 0.4 & 89.98 $\pm$ 0.5 & 91.58 $\pm$ 0.2 & 96.53 $\pm$ 0.1 & 95.51 $\pm$ 0.2 & 97.57 $\pm$ 1.3 & 93.96 $\pm$ 0.2 & 93.32 $\pm$ 0.5 & 97.01 $\pm$ 0.6\\
M3D-CNN & 95.32 $\pm$ 0.1 & 94.70 $\pm$ 0.2 & 96.41 $\pm$ 0.7 & 95.76 $\pm$ 0.2 & 94.50 $\pm$ 0.2 & 95.08 $\pm$ 1.2 & 94.79 $\pm$ 0.3 & 94.20 $\pm$ 0.2 & 96.25 $\pm$ 0.6\\
SSRN & 99.19 $\pm$ 0.3 & 99.07 $\pm$ 0.3 & 98.93 $\pm$ 0.6 & 99.90 $\pm$ 0.0 & 99.87 $\pm$ 0.0 & 99.91 $\pm$ 0.0 & 99.98 $\pm$ 0.1 & 99.97 $\pm$ 0.1 & 99.97 $\pm$ 0.0\\
\textbf{HybridSN} & 99.75 $\pm$ 0.1 & 99.71 $\pm$ 0.1 & 99.63 $\pm$ 0.2 & 99.98 $\pm$ 0.0 & 99.98 $\pm$ 0.0 & 99.97 $\pm$ 0.0 & 100 $\pm$ 0.0 & 100 $\pm$ 0.0 & 100 $\pm$ 0.0\\
\bottomrule
\end{tabular}
\label{tab:comp}
\end{table*}

In 2D-CNN, the input data are convolved with 2D kernels. The convolution happens by computing the sum of the dot product between input data and kernel. The kernel is strided over the input data to cover full spatial dimension. The convolved features are passed through the activation function to introduce the non-linearity in the model. In 2D convolution, the activation value at spatial position $(x,y)$ in the $j^{th}$ feature map of the $i^{th}$ layer, denoted as $v^{x,y}_{i,j}$, is generated using the following equation,
    \begin{equation}
    \label{equ:2D-cnn}
    \begin{aligned}
    v^{x,y}_{i,j} = \phi(b_{i,j} + \sum_{\tau=1}^{d_{l-1}}\sum_{\rho=-\gamma}^{\gamma}\sum_{\sigma=-\delta}^{\delta}w_{i,j,\tau}^{\sigma, \rho} \times v_{i-1,\tau}^{x+\sigma,y+\rho})
    \end{aligned}    
    \end{equation}
where $\phi$ is the activation function, $b_{i,j}$ is the bias parameter for the $j^{th}$ feature map of the $i^{th}$ layer, $d_{l-1}$ is the number of feature map in $(l-1)^{th}$ layer and the depth of kernel $w_{i,j}$ for the $j^{th}$ feature map of the $i^{th}$ layer, $2\gamma+1$ is the width of kernel, $2\delta+1$ is the height of kernel, and $w_{i,j}$ is the value of weight parameter for the $j^{th}$ feature map of the $i^{th}$ layer.

The 3D convolution~\cite{ji20133d} is done by convolving a 3D kernel with the 3D-data. In the proposed model for HSI data, the feature maps of convolution layer are generated using the 3D kernel over multiple contiguous bands in the input layer; this captures the spectral information. In 3D convolution, the activation value at spatial position $(x,y,z)$ in the $j^{th}$ feature map of the $i^{th}$ layer, denoted as $v^{x,y,z}_{i,j}$, is generated as follows,
\begin{equation}
\label{equ:3D-cnn}
\begin{aligned}
    v^{x,y,z}_{i,j} = \phi(b_{i,j} + \sum_{\tau=1}^{d_{l-1}}\sum_{\lambda=-\eta}^{\eta}\sum_{\rho=-\gamma}^{\gamma}\sum_{\sigma=-\delta}^{\delta}w_{i,j,\tau}^{\sigma, \rho,\lambda} \times v_{i-1,\tau}^{x+\sigma,y+\rho,z+\lambda})
\end{aligned}    
\end{equation}
where $2\eta+1$ is the depth of kernel along spectral dimension and other parameters are the same as in (Eqn. 1).

The parameters of CNN, such as the bias $b$ and the kernel weight $w$, are usually trained using supervised approaches \cite{krizhevsky2012imagenet} with the help of a gradient descent optimization technique. In conventional 2D CNNs, the convolutions are applied over the spatial dimensions only, covering all the feature maps of the previous layer, to compute the 2D discriminative feature maps. Whereas, for the HSI classification problem, it is desirable to capture the spectral information, encoded in multiple bands along with the spatial information. The 2D-CNNs are not able to handle the spectral information. On the other hand, the 3D-CNN kernel can extract the spectral and spatial feature representation simultaneously from HSI data, but at the cost of increased computational complexity. 
In order to take the advantages of the automatic feature learning capability of both 2D and 3D CNN, we propose a hybrid feature learning framework called $HybridSN$ for HSI classification. The flow diagram of the proposed $HybridSN$ network is shown in Fig.~\ref{fig:model}.
It comprises of three 3D convolutions (Eqn.~\ref{equ:3D-cnn}), one 2D convolution (Eqn.~\ref{equ:2D-cnn}) and three fully connected layers.

In $HybridSN$ framework, the dimensions of 3D convolution kernels are $8\times{3}\times{3}\times{7}\times{1}$ (i.e., $K_1^1=3$, $K_2^1=3$, and $K_3^1=7$ in Fig. \ref{fig:model}), $16\times{3}\times{3}\times{5}\times{8}$ (i.e., $K_1^2=3$, $K_2^2=3$, and $K_3^2=5$ in Fig. \ref{fig:model}) and $32\times{3}\times{3}\times{3}\times{16}$ (i.e., $K_1^3=3$, $K_2^3=3$, and $K_3^3=3$ in Fig. \ref{fig:model}) in the subsequent $1^{st}$, $2^{nd}$ and $3^{rd}$ convolution layers, respectively, where $16\times{3}\times{3}\times{5}\times{8}$ means $16$ 3D-kernels of dimension ${3}\times{3}\times{5}$ (i.e., two spatial and one spectral dimension) for all $8$ 3D input feature maps. Whereas, the dimension of 2D convolution kernel is $64\times{3}\times{3}\times{576}$ (i.e., $K_1^4=3$ and $K_2^4=3$ in Fig. \ref{fig:model}), where $64$ is the number of 2D-kernels, ${3}\times{3}$ represents the spatial dimension of 2D-kernel, and $576$ is the number of 2D input feature maps. 
To increase the number of spectral-spatial feature maps simultaneously, 3D convolutions are applied thrice and can preserve the spectral information of input HSI data in the output volume. The 2D convolution is applied once before the $flatten$ layer by keeping in mind that it strongly discriminates the spatial information within the different spectral bands without substantial loss of spectral information, which is very important for HSI data. A detailed summary of the proposed model in terms of the layer types, output map dimensions and number of parameters is given in Table \ref{tab:model_summ}. It can be seen that the highest number of parameters are present in the $1^{st}$ dense layer. The number of node in the last dense layer is 16, which is same as the number of classes in Indian Pines dataset. Thus, the total number of parameters in the proposed model depends on the number of classes in a dataset. The total number of trainable weight parameters in $HybridSN$ is $5,122,176$ for Indian Pines dataset. All weights are randomly initialised and trained using back-propagation algorithm with the $Adam$ optimiser by using the $softmax$ loss. We use mini-batches of size $256$ and train the  network for $100$ epochs with no batch normalization and data augmentation.

\begin{figure*}[!t]
\begin{subfigure}{.255\columnwidth}
\centering
\includegraphics[clip=true, trim = 0 0 0 0, width=0.98\columnwidth]{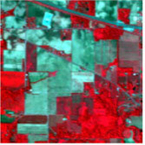}
\caption{}
\end{subfigure}%
\begin{subfigure}{.255\columnwidth}
\centering
\includegraphics[clip=true, trim = 0 0 0 0, width=0.98\columnwidth]{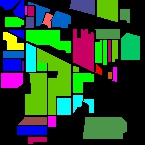}
\caption{}
\end{subfigure}%
\begin{subfigure}{.255\columnwidth}
\centering
\includegraphics[clip=true, trim = 0 0 0 0, width=0.98\columnwidth]{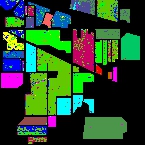}
\caption{}
\end{subfigure}%
\begin{subfigure}{.255\columnwidth}
\centering
\includegraphics[clip=true, trim = 0 0 0 0, width=0.98\columnwidth]{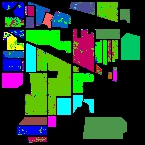}
\caption{}
\end{subfigure}%
\begin{subfigure}{.255\columnwidth}
\centering
\includegraphics[clip=true, trim = 0 0 0 0, width=0.98\columnwidth]{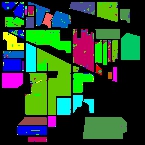}
\caption{}
\end{subfigure}%
\begin{subfigure}{.255\columnwidth}
\centering
\includegraphics[clip=true, trim = 0 0 0 0, width=0.98\columnwidth]{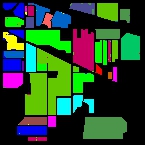}
\caption{}
\end{subfigure}%
\begin{subfigure}{.255\columnwidth}
\centering
\includegraphics[clip=true, trim = 0 0 0 0, width=0.98\columnwidth]{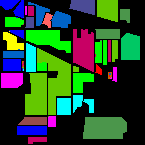}
\caption{}
\end{subfigure}%
\begin{subfigure}{.255\columnwidth}
\centering
\includegraphics[clip=true, trim = 0 0 0 0, width=0.98\columnwidth]{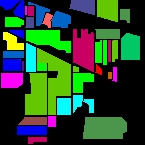}
\caption{}
\end{subfigure}
\caption{The classification map for Indian Pines, (a) False color image (b) Ground truth (c)-(h) Predicted classification maps for SVM, 2D-CNN, 3D-CNN, M3D-CNN, SSRN, and proposed HybridSN, respectively.}
\label{fig:map_IP}
\end{figure*}

\section{Experiments and Discussion}

\subsection{Dataset Description and Training Details}
We have used three publicly available hyperspectral image datasets\footnote{www.ehu.eus/ccwintco/index.php/Hyperspectral\_Remote\_Sensing\_Scenes}, namely Indian Pines, University of Pavia and Salinas Scene. The \textbf{Indian Pines (IP) dataset} has images with $145\times{145}$ spatial dimension and $224$ spectral bands in the wavelength range of $400$ to $2500$ $nm$, out of which $24$ spectral bands covering the region of water absorption have been discarded. The ground truth available is designated into $16$ classes of vegetation. The \textbf{University of Pavia (UP) dataset} consists of $610\times{340}$ spatial dimension pixels with $103$ spectral bands in the wavelength range of $430$ to $860~nm$. The ground truth is divided into $9$ urban land-cover classes. The \textbf{Salinas Scene (SA) dataset} contains the images with $512\times{217}$ spatial dimension and $224$ spectral bands in the wavelength range of $360$ to $2500~nm$. The $20$ water absorbing spectral bands have been discarded. In total $16$ classes are present in this dataset.

All experiments are conducted on an Acer Predator-Helios laptop with the GTX 1060 Graphical Processing Unit (GPU) and 16 GB of RAM. We have chosen the optimal learning rate of $0.001$, based on the classification outcomes. In order to make the fair comparison, we have extracted the same spatial dimension in 3D-patches of input volume for different datasets, such as $25\times{25}\times{30}$ for IP and $25\times{25}\times{15}$ for UP and SA, respectively.

\begin{figure}[!t]
\centering
\begin{subfigure}{0.32\columnwidth}
\centering
\includegraphics[clip=true, trim = 130 290 150 315, width=\textwidth]{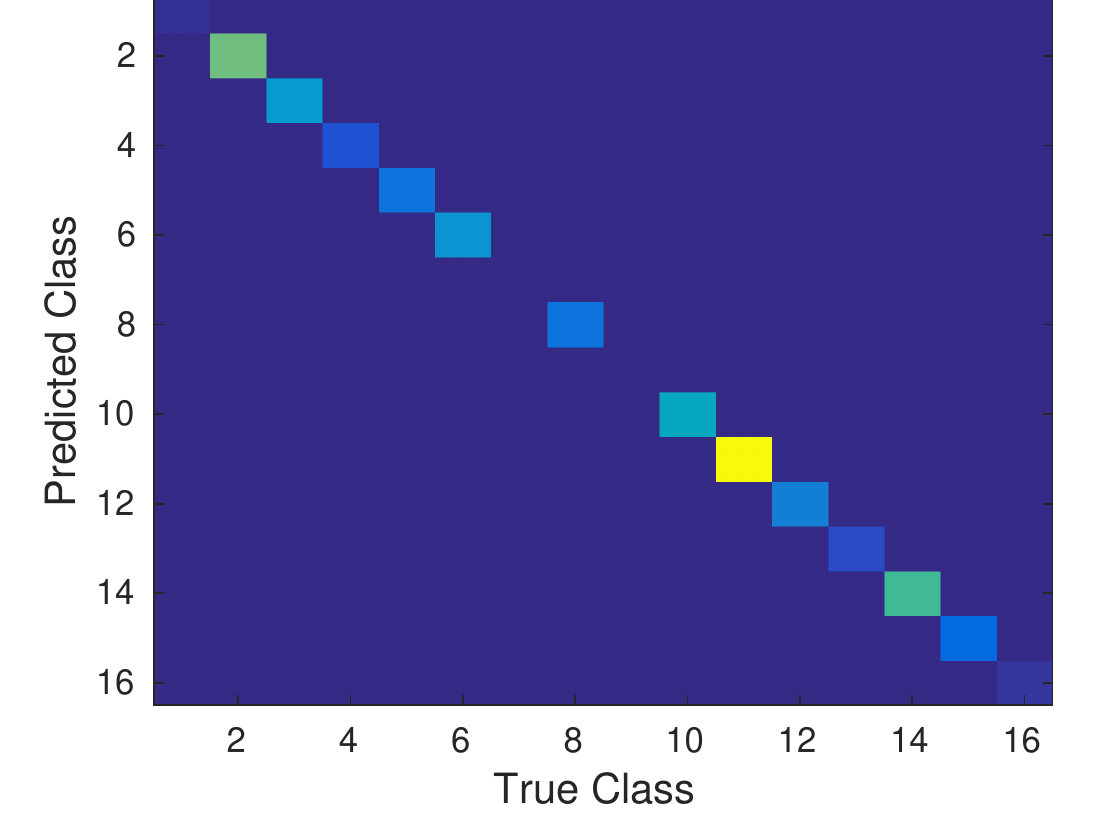}
\end{subfigure}%
\begin{subfigure}{0.32\columnwidth}
\centering
\includegraphics[clip=true, trim = 130 290 150 315, width=\columnwidth]{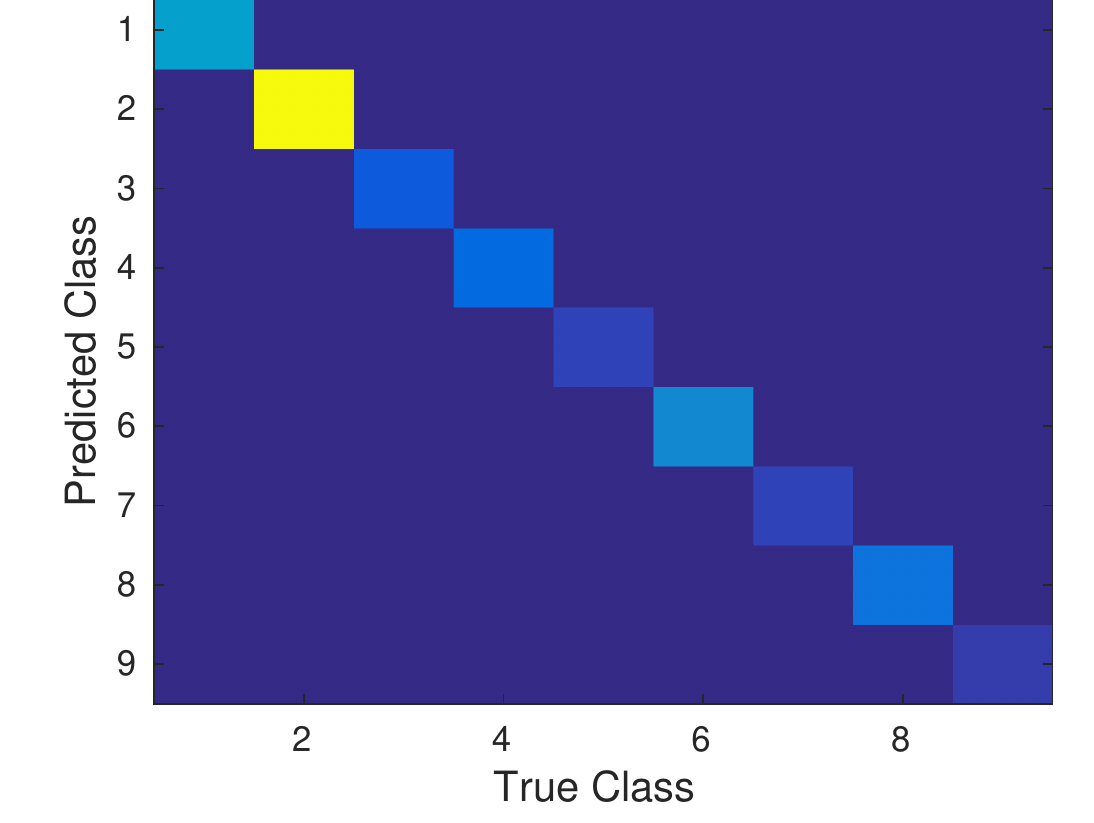}
\end{subfigure}
\begin{subfigure}{0.32\columnwidth}
\centering
\includegraphics[clip=true, trim = 130 290 150 315, width=\columnwidth]{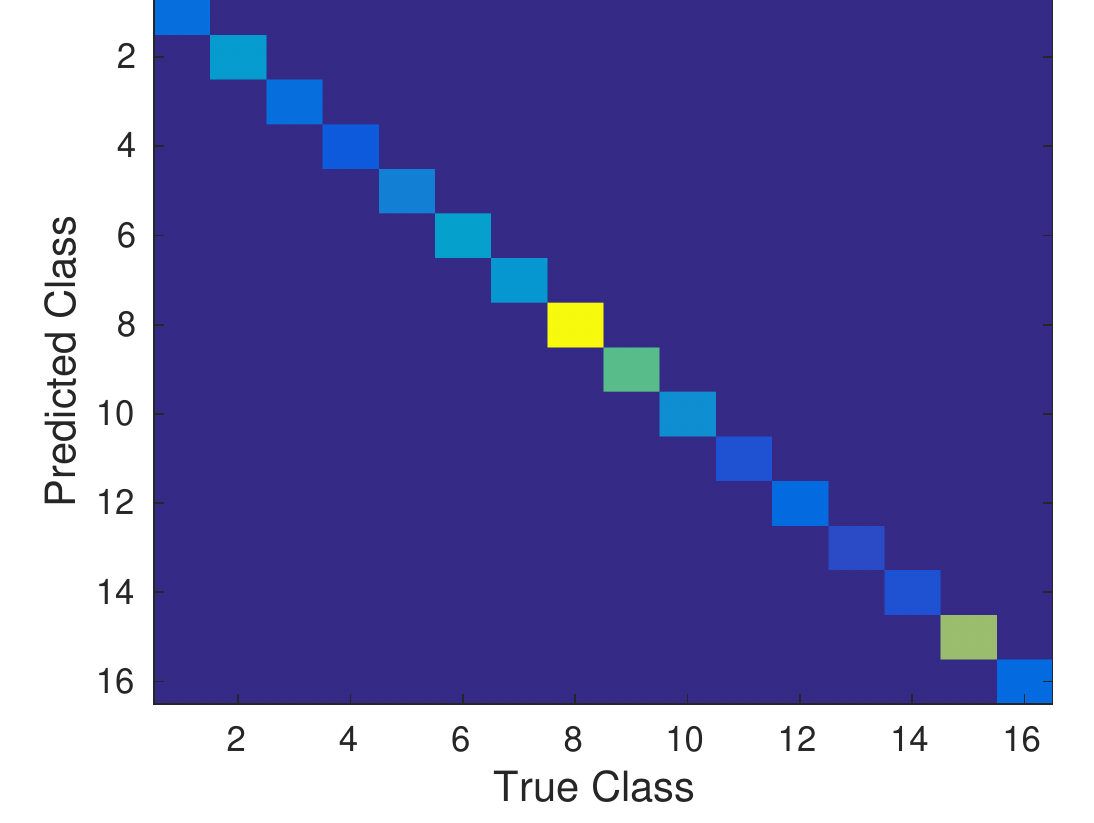}
\end{subfigure}%
\caption{The confusion matrix using proposed method over Indian Pines, University of Pavia, and Salinas Scene datasets in $1^{st}$, $2^{nd}$, and $3^{rd}$ matrix, respectively.}
\label{fig:conf_mat}
\end{figure}

\begin{figure}[!t]
\centering
\includegraphics[clip=true, trim = 00 280 50 05, width=0.99\columnwidth, height=30mm]{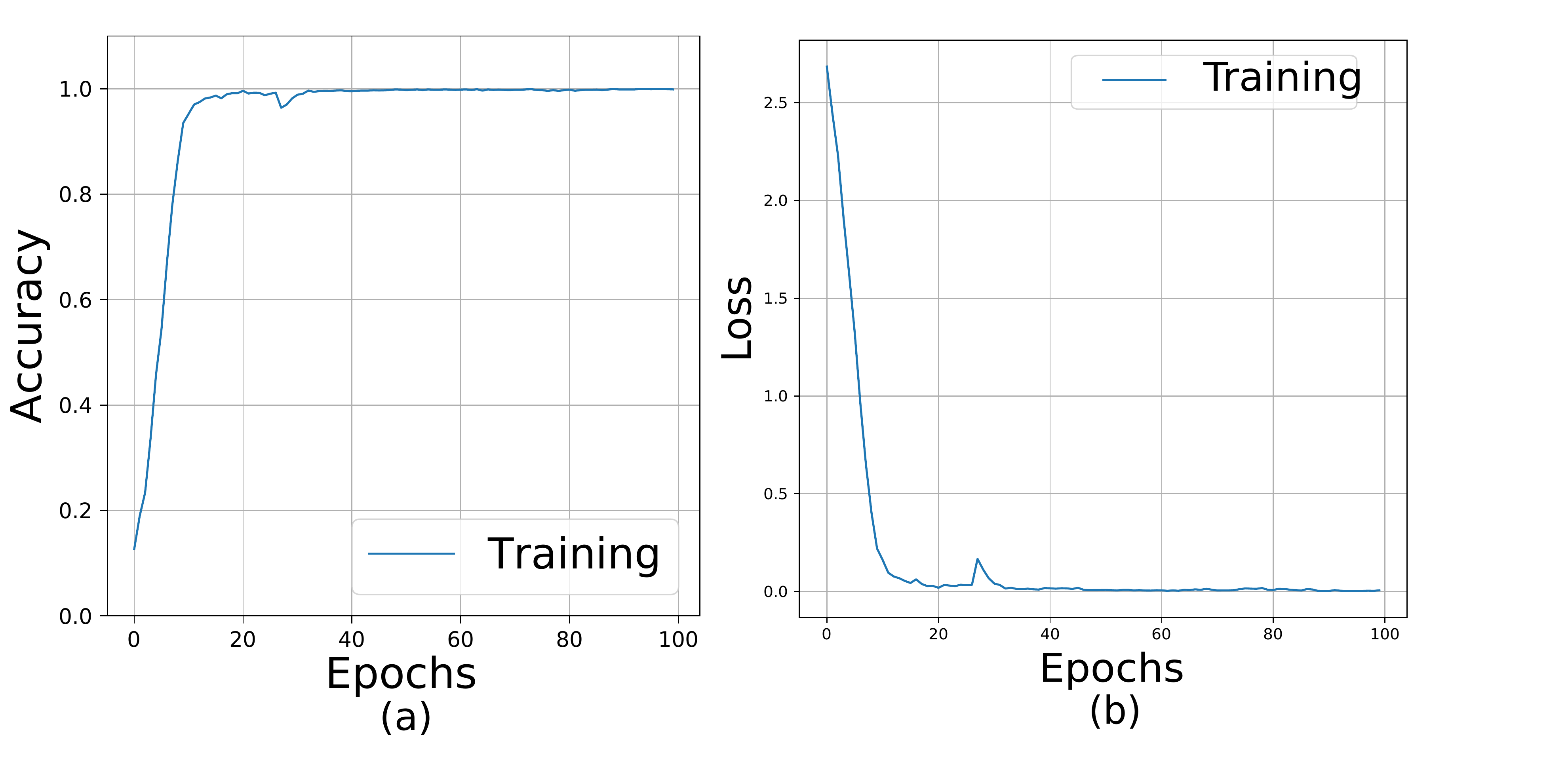}
\caption{The accuracy and loss convergence vs epochs over Indian Pines dataset.}
\label{fig:acc_loss}
\end{figure}


\subsection{Classification Results}
In this letter, we have used the Overall Accuracy (OA), Average Accuracy (AA) and Kappa Coefficient (Kappa) evaluation measures to judge the HSI classification performance. Here, OA represents the number of correctly classified samples out of the total test samples; AA represents the average of class-wise classification accuracies; and Kappa is a metric of statistical measurement which provides mutual information regarding a strong agreement between the ground truth map and classification map. The results of the  proposed $HybridSN$ model are compared with the most widely used supervised methods, such as SVM~\cite{melgani2004classification}, 2D-CNN~\cite{makantasis2015deep}, 3D-CNN~\cite{hamida2018deep}, M3D-CNN~\cite{he2017multi}, and SSRN~\cite{zhong2018spectral}. The 30\% and 70\% of the data are randomly divided into training and testing groups, respectively\footnote{More details of dataset are provided in the supplementary material}. We have used the publicly available code\footnote{\url{https://github.com/eecn/Hyperspectral-Classification}} of the compared methods to compute the results.

Table~\ref{tab:comp} shows the results in terms of the OA, AA, and Kappa coefficient for different methods\footnote{The class-wise accuracy is provided in the supplementary material.}. It can be observed from Table~\ref{tab:comp} that the $HybridSN$ outperforms all the compared methods over each dataset while maintaining the minimum standard deviation. The proposed $HybridSN$ is based on the hierarchical representation of spectral-spatial 3D CNN followed by a spatial 2D CNN, which are complementary to each other. 
It is also observed from these results that the performance of 3D-CNN is poor than 2D-CNN over Salinas Scene dataset. To the best of the our knowledge, this is probably due to the presence of two classes in the Salinas dataset (namely Grapes-untrained and Vinyard-untrained) which have very similar textures over most spectral bands. Hence, due to the increased redundancy among the spectral bands, the 2D-CNN outperforms the 3D-CNN over Salinas Scene dataset.
Moreover, the performance of SSRN and $HybridSN$ is always far better than M3D-CNN. It is evident that 3D or 2D convolution alone is not able to represent the highly discriminative feature compared to hybrid 3D and 2D convolutions.

\begin{table}[!t]
\caption{The training time in minutes (m) and test time in seconds (s) over IP, UP, and SA datasets using 2D-CNN, 3D-CNN and $HybridSN$ architectures.}
\centering
\begin{tabular}{c|c|c|c|c|c|c}
\toprule
\multirow{2}{*}{Data} & \multicolumn{2}{c|}{2D CNN}
& \multicolumn{2}{c|}{3D CNN}
& \multicolumn{2}{c}{HybridSN} \\ 
\cline{2-7} 
& Train(m) & Test(s) & Train(m) & Test(s) & Train(m) & Test(s) \\ 
\hline
IP & 1.9 & 1.1 & 15.2 & 4.3 & 14.1 & 4.8 \\
UP & 1.8 & 1.3 & 58.0 & 10.6 & 20.3 & 6.6 \\
SA & 2.2 & 2.0 & 74 & 15.2 & 25.5 & 9.0 \\ 
\bottomrule
\end{tabular}
\label{tab:time}
\end{table}

\begin{table}[!t]
\centering
\caption{The impact of spatial window size over the performance of $HybridSN$.}
\begin{tabular}{m{0.7cm}m{0.65cm}m{0.65cm}m{0.65cm}|m{0.7cm}m{0.65cm}m{0.65cm}m{0.65cm}}
\toprule
Window & IP(\%) & UP(\%) & SA(\%) & Window & IP(\%) & UP(\%) & SA(\%)\\
\midrule
19$\times$19 & 99.74 & 99.98 & 99.99 & 23$\times$23 & 99.31 & 99.96 & 99.71\\
21$\times$21 & 99.73 & 99.90 & 99.69 & 25$\times$25 & 99.75 & 99.98 & 100\\ 
\bottomrule
\end{tabular}
\label{tab:spatial}
\end{table}

\begin{table}[!t]
\caption{The classification accuracies (in percentages) using proposed and state-of-the-art methods on less amount of training data, i.e., 10\% only.}
\centering
\begin{tabular}{m{1.29cm}|m{0.3cm}m{0.4cm}m{0.4cm}|m{0.3cm}m{0.4cm}m{0.4cm}|m{0.3cm}m{0.4cm}m{0.4cm}}
\toprule
\multirow{2}{*}{Methods} & \multicolumn{3}{m{1.5cm}|}{Indian Pines} & \multicolumn{3}{m{1.7cm}|}{Univ. of Pavia} & \multicolumn{3}{m{1.6cm}}{Salinas Scene}\\
\cline{2-10}
 & OA & Kappa & AA & OA & Kappa & AA & OA & Kappa & AA\\
\midrule
2D-CNN & 80.27 & 78.26 & 68.32 & 96.63 & 95.53 & 94.84 & 96.34 & 95.93 & 94.36\\
3D-CNN & 82.62 & 79.25 & 76.51 & 96.34 & 94.90 & 97.03 & 85.00 & 83.20 & 89.63\\
M3D-CNN & 81.39 & 81.20 & 75.22 & 95.95 & 93.40 & 97.52 & 94.20 & 93.61 & 96.66\\
SSRN & 98.45 & 98.23 & 86.19 & 99.62 & 99.50 & 99.49 & 99.64 & 99.60 & 99.76\\
\textbf{HybridSN} & 98.39 & 98.16 & 98.01 & 99.72 & 99.64 & 99.20 & 99.98 & 99.98 & 99.98\\
\bottomrule
\end{tabular}
\label{tab:lesstraining}
\end{table}

The classification map for an example hyperspectral image is illustrated in Fig.~\ref{fig:map_IP} using SVM, 2D-CNN, 3D-CNN, M3D-CNN, SSRN and $HybridSN$ methods. The quality of classification map of SSRN and $HybridSN$ is far better than other methods. Among SSRN and HybridSN, the maps generated by $HybridSN$ in small segment are better than SSRN.
Fig.~\ref{fig:conf_mat} shows the confusion matrix for the HSI classification performance of the proposed $HybridSN$ over IP, UP and SA datasets, respectively. The accuracy and loss convergence for 100 epochs of training and validation sets are portrayed in Fig.~\ref{fig:acc_loss} for the proposed method. It can be seen that the convergence is achieved in approximately 50 epochs which points out the fast convergence of our method. The computational efficiency of $HybridSN$ model appears in term of training and testing times in Table~\ref{tab:time}. The proposed model is more efficient than 3D-CNN. The impact of spatial dimension over the performance of $HybridSN$ model is reported in Table \ref{tab:spatial}. It has been found that the used $25\times25$ spatial dimension is most suitable for the proposed method. We have also computed the results with an even less training data, i.e., only $10\%$ of total samples and have summarized the results in Table \ref{tab:lesstraining}. It is observed from this experiment that the performance of each model decreases slightly, whereas the proposed method is still able to outperform other methods in almost all cases.

\section{Conclusion}
This letter has introduced a hybrid 3D and 2D model for hyperspectral image classification. The proposed $HybridSN$ model basically combines the complementary information of spatio-spectral and spectral in the form of 3D and 2D convolutions, respectively. The experiments over three benchmark datasets compared with recent state-of-the-art methods confirm the superiority of the proposed method. The proposed model is computationally efficient than the 3D-CNN model. It also shows the superior performance for small training data.

\bibliographystyle{IEEEtran}
\bibliography{Reference}
\end{document}